\pdfoutput=1
\documentclass[11pt]{article}

\usepackage{acl}

\usepackage{times}
\usepackage{latexsym}

\usepackage[T1]{fontenc}
\usepackage[utf8]{inputenc}

\usepackage{microtype}

\usepackage{inconsolata}

\usepackage{array}
\usepackage{multirow}
\usepackage{footmisc}  % Package to allow referencing the same footnote multiple times
\usepackage{float}
\usepackage{hyperref}
\usepackage{amsmath, amssymb}

\usepackage{tcolorbox}
\newtcbox{\bluebox}{colback=cyan, boxrule=0pt, arc=5pt, boxsep=0pt, left=3pt, right=3pt, top=3pt, bottom=3pt, nobeforeafter, height=7mm, valign=center, baseline=2.5mm}
\newtcbox{\greybox}{colback=lightgray, boxrule=0pt, arc=5pt, boxsep=0pt, left=3pt, right=3pt, top=3pt, bottom=3pt, nobeforeafter, height=7mm, valign=center, baseline=2.5mm}
\newtcbox{\redbox}{colback=purple, boxrule=0pt, arc=5pt, boxsep=0pt, left=3pt, right=3pt, top=3pt, bottom=3pt, nobeforeafter, height=7mm, valign=center, baseline=2.5mm}

\title{Soft Prompt Tuning for Cross-Lingual Transfer: When Less is More}

\author{Fred Philippy\textsuperscript{$1,2$}, Siwen Guo\textsuperscript{$1$}, Shohreh Haddadan\textsuperscript{$1$},\vspace{0.2cm} \\ \textbf{Cedric Lothritz\textsuperscript{$2$}, Jacques Klein\textsuperscript{$2$}, Tegawendé F. Bissyandé\textsuperscript{$2$}} \vspace{0.4cm} \\ \textsuperscript{$1$} Zortify S.A., Luxembourg \\ \textsuperscript{$2$} University of Luxembourg, Luxembourg \\ \texttt{\{fred, siwen\}@zortify.com, shohreh.haddadan@gmail.com} \\ \texttt{\{cedric.lothritz, jacques.klein, tegawende.bissyande\}@uni.lu}}

\begin{document}
\maketitle

\begin{abstract}
Soft Prompt Tuning (SPT) is a parameter-efficient method for adapting pre-trained language models (PLMs) to specific tasks by inserting learnable embeddings, or soft prompts, at the input layer of the PLM, without modifying its parameters. This paper investigates the potential of SPT for cross-lingual transfer. Unlike previous studies on SPT for cross-lingual transfer that often fine-tune both the soft prompt and the model parameters, we adhere to the original intent of SPT by keeping the model parameters frozen and only training the soft prompt. This does not only reduce the computational cost and storage overhead of full-model fine-tuning, but we also demonstrate that this very parameter efficiency intrinsic to SPT can enhance cross-lingual transfer performance to linguistically distant languages. Moreover, we explore how different factors related to the prompt, such as the length or its reparameterization, affect cross-lingual transfer performance.
\end{abstract}
\section{Introduction}
Fine-tuning pre-trained language models (PLMs) on task-specific labeled data requires large amounts of computational resources and may cause catastrophic forgetting of the pre-trained knowledge \citep{goodfellow_empirical_2015}. In multilingual settings, this may lead to poor cross-lingual transfer performance \citep{vu_overcoming_2022}.

To address these challenges, \citet{lester_power_2021} introduced Soft Prompt Tuning (SPT), a method that inserts learnable embeddings, or soft prompts, at the PLM's input layer. The PLM then makes predictions using the output of its pre-trained language modeling head. The key advantage of SPT lies in its ability to leverage the pre-existing knowledge within PLMs while reducing the reliance on extensive task-specific fine-tuning. SPT has been shown to achieve remarkable results in various monolingual downstream tasks, especially in few-shot settings.

Motivated by this success, some recent works have also explored the use of SPT for cross-lingual transfer, where the goal is to leverage a multilingual language model (MLLM) to transfer knowledge from a high-resource to a low-resource language. However, these works have not fully exploited the potential of SPT. Some have appended a newly initialized classifier to the model \citep{tu_prompt-tuning_2022, park_analysis_2023}, hindering the suitability of SPT for few-shot learning. Others have fine-tuned the entire model along with the prompt \citep{zhao_discrete_2021, huang_zero-shot_2022}, which reduces the computational efficiency of SPT.

This is especially problematic given the growing size of state-of-the-art language models. Therefore, we explore the impact on SPT's cross-lingual transfer performance when adhering to the original methodology of \citet{lester_power_2021}, which involves fine-tuning only the soft prompt while keeping all model parameters frozen. Specifically, this paper contributes to the field of cross-lingual SPT by:
\begin{itemize}
  \setlength{\itemsep}{1pt}
  \setlength{\parskip}{0pt}
  \setlength{\parsep}{0pt}
    \item Investigating the impact of model freezing on the cross-lingual transfer performance of few-shot SPT.
    \item Demonstrating that by freezing the model, SPT achieves enhanced cross-lingual transfer, especially to languages linguistically distant from the source language.
    \item Exploring further non-linguistic factors that influence the cross-lingual transfer performance of SPT, in particular prompt length and prompt reparameterization.
\end{itemize}

In this study, we conduct experiments on a topic classification dataset in 52 different languages and using 4 different models in few-shot settings.
We believe that our findings can improve the existing methods that aim to enhance cross-lingual SPT, particularly in the context of current state-of-the-art models with billions of parameters where parameter efficiency is crucial. 

\section{Related Work}
\citet{lester_power_2021} proposed SPT, a method to leverage a PLM's pre-trained language modeling head without appending a new classifier. SPT relies on a soft prompt, which is a set of learnable embeddings that are concatenated with the input sequence, and keeps all other model parameters frozen.
Since then, several recent works have explored the use of soft prompts for MLLMs. \citet{zhao_discrete_2021} first show that SPT outperforms fine-tuning in few-shot scenarios for cross-lingual transfer. 
\citet{huang_zero-shot_2022} introduce a method to train a language-agnostic soft prompt.
However, unlike our study, none of these works on cross-lingual SPT employ model parameter freezing, leading to a reduced efficiency in their methods. 
In contrast, \citet{tu_prompt-tuning_2022} and \citet{park_analysis_2023} perform model freezing and, in corroboration with \citet{zhao_discrete_2021}, also show that SPT outperforms fine-tuning for cross-lingual transfer. 
However, they append a newly initialized classification head to the model instead of using the PLM's pre-trained language modeling head, which diverges from the original idea of SPT. 
This setup is unsuitable for few-shot learning, requiring experiments to be conducted in full-data settings.
In addition, prior studies often focus on smaller ranges of languages, which impedes making conclusive observations about SPT's cross-lingual tendencies across different languages and language families.

\section{Experimental Setup} \label{sec:experimental_setup}
Besides adhering to the original setup of SPT, enabling parameter-efficient and data-efficient training, our study also sets itself apart in its objectives from the existing literature. 
Rather than simply demonstrating superior cross-lingual transfer performance of SPT over fine-tuning, our research aims to show that the minimal impact on the MLLM's representation space not only generally enhances transfer performance but is particularly effective for linguistically distant languages.

We provide more specific details on our experimental setup in Appendix \ref{app:reproducibility}.

\subsection{Soft Prompt} \label{sec:soft_prompt}
\begin{figure}[!h]
    \centering
    \includegraphics[width=0.45\textwidth]{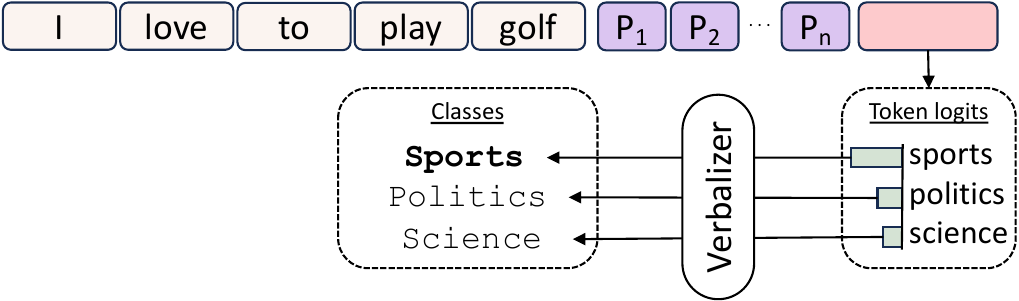}
    \caption{A simplified illustration of SPT \citep{lester_power_2021}. $P_1, \ldots, P_n$ denote the soft prompt tokens, with each token corresponding to a trainable embedding. Essentially, for a model with an embedding dimension $d$, a soft prompt of length $n$ forms a $d \times n$ matrix.}
    \label{fig:soft_prompt}
\end{figure}

Following \citet{lester_power_2021}, we append a soft prompt to the input sequence which is passed through an autoregressive language model, generating the logits for the next token in the input sequence. Each class is linked to a token from the model's vocabulary, enabling us to map the token with the highest logit to the predicted class. Such a mapping is referred to as the \textit{verbalizer} (Figure \ref{fig:soft_prompt}).

\subsection{Implementation Details} \label{implementation}
\paragraph{Models}
With the recent advancement and popularity of autoregressive language models for various tasks, our research is conducted using two types of MLLMs based on this architecture: XGLM \citep{lin_few-shot_2022} and BLOOM \citep{scao_bloom_2022}. For both models we use 2 different sizes: XGLM-564M and XGLM-1.7B for XGLM, and BLOOM-560M and BLOOM-1.1B for BLOOM.  

\paragraph{Data}
In our study, we use SIB-200 \citep{adelani_sib-200_2023}, a topic classification dataset containing seven distinct topics and covering a diverse range of 200 languages and dialects.
We chose this dataset for its broader, more diverse language range compared to prior studies on cross-lingual SPT, covering almost all languages our models support, enabling more comprehensive observations.

\paragraph{Technical Details}
We compare two different settings: tuning the soft prompt with model freezing (\textit{w/ MF}) and without model freezing (\textit{w/o MF}). We perform few-shot fine-tuning only using English samples.
The final cross-lingual transfer performance is then evaluated on the test sets of all languages supported by the respective model (30 for XGLM, 38 for BLOOM), using accuracy as the metric. We repeat each experiment 4 times with different random seeds and report the mean.

\section{Results} \label{sec:results}

\begin{table*}[ht]
\small
\centering
\setlength{\tabcolsep}{8pt}
\renewcommand{\arraystretch}{1.5}
\begin{tabular}{cc|ccccccc}
&  & \textbf{DATA} & \textbf{SYN} & \textbf{GEO} & \textbf{INV} & \textbf{GEN} & \textbf{PHON} & \textbf{FEA} \\ \hline
\multirow{2}{*}{\textit{\textbf{BLOOM-560M}}} & \textit{w/o MF} & 0,6781 & 0,6457 & 0,2294 & 0,3779 & 0,5081 & \textbf{0,4343} & 0,4221 \\
 & \textit{w/ MF} & \textbf{0,6080} & \textbf{0,5742} & \textbf{0,2034} & \textbf{0,2629} & \textbf{0,3676} & 0,4482 & \textbf{0,3165} \\ \hline
\multirow{2}{*}{\textit{\textbf{BLOOM-1.1B}}} & \textit{w/o MF} & 0,6788 & 0,6403 & 0,1693 & 0,4605 & 0,5679 & 0,5272 & -0,4685 \\
 & \textit{w/ MF} & \textbf{0,4856} & \textbf{0,4177} & \textbf{0,0290} & \textbf{0,2930} & \textbf{0,3711} & \textbf{0,4283} & \textbf{0,3002} \\ \hline
\multirow{2}{*}{\textit{\textbf{XGLM-564M}}} & \textit{w/o MF} & 0,2672 & 0,6767 & 0,4694 & 0,4016 & 0,3203 & 0,4756 & 0,5949 \\
 & \textit{w/ MF} & \textbf{0,2453} & \textbf{0,6574} & \textbf{0,2551} & \textbf{0,3410} & \textbf{0,2201} & \textbf{0,3285} & \textbf{0,5185} \\ \hline
\multirow{2}{*}{\textit{\textbf{XGLM-1.7B}}} & \textit{w/o MF} & 0,2636 & 0,6722 & \textbf{0,2566} & 0,3623 & 0,2924 & 0,3213 & 0,5315 \\
 & \textit{w/ MF} & \textbf{0,2560} & \textbf{0,6694} & 0,2949 & \textbf{0,3155} & \textbf{0,2786} & \textbf{0,2779} & \textbf{0,4922} \\ \hline
\end{tabular}
\caption{Pearson correlation between (8-shot) cross-lingual transfer performance and 6 different linguistic similarity metrics, namely syntactic (SYN), geographic (GEO), inventory (INV), genetic (GEN), phonological (PHON) and featural (FEA) distance, as well as the language-specific pre-training corpus size (DATA).}
\label{table:correlation}
\end{table*}

\begin{figure}
    \centering
    \includegraphics[width=0.45\textwidth]{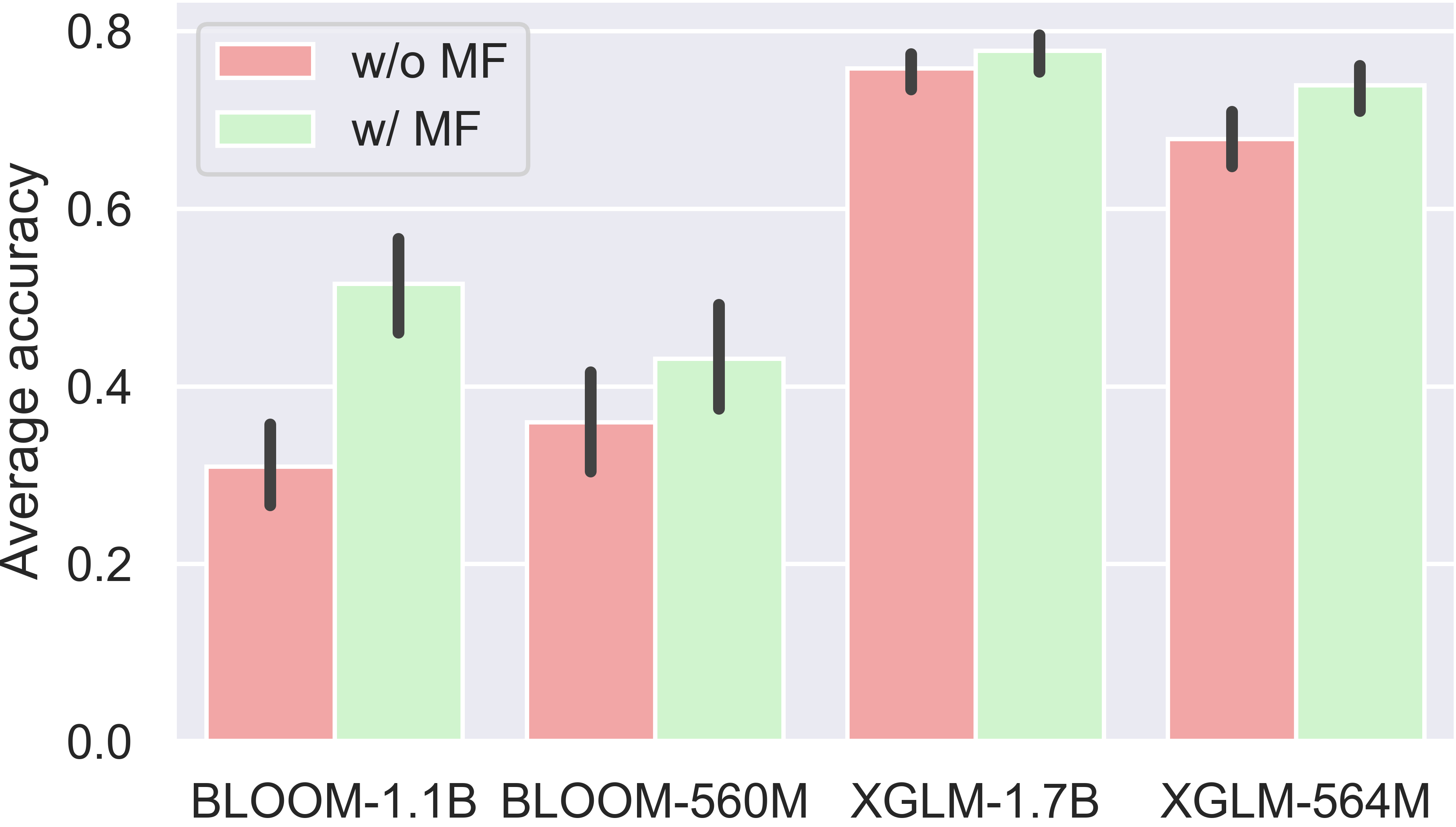}
    \caption{Average cross-lingual transfer performance of SPT with and without model freezing (MF) for different models across all languages supported by the respective model.}
    \label{fig:base_results}
\end{figure}

We provide the full results across all models and languages in Appendix \ref{app:results}.
The results reveal that model freezing not only \textbf{boosts cross-lingual transfer performance} (Figure \ref{fig:base_results}) but additionally is a step towards \textbf{closing the transfer gap} between linguistically distant and similar languages. This can be seen in Table \ref{table:correlation}, which shows that the correlation strength between transfer performance and language similarity between source and target languages, measured using 6 different similarity metrics\footnote{See Appendix \ref{app:lang_dist} for more details.} \citep{littell_uriel_2017}, decreases when freezing model parameters. 
This suggests that the parameter efficiency of SPT mitigates the bias of cross-lingual transfer towards linguistically similar languages. In other words, by fine-tuning fewer parameters, cross-lingual transfer, especially to linguistically distant languages, is enhanced.
This improvement over full-model fine-tuning may be attributed to the reduced impact on the MLLM's representation space during fine-tuning \citep{philippy_identifying_2023}.

\begin{figure}
    \centering
    \includegraphics[width=0.45\textwidth]{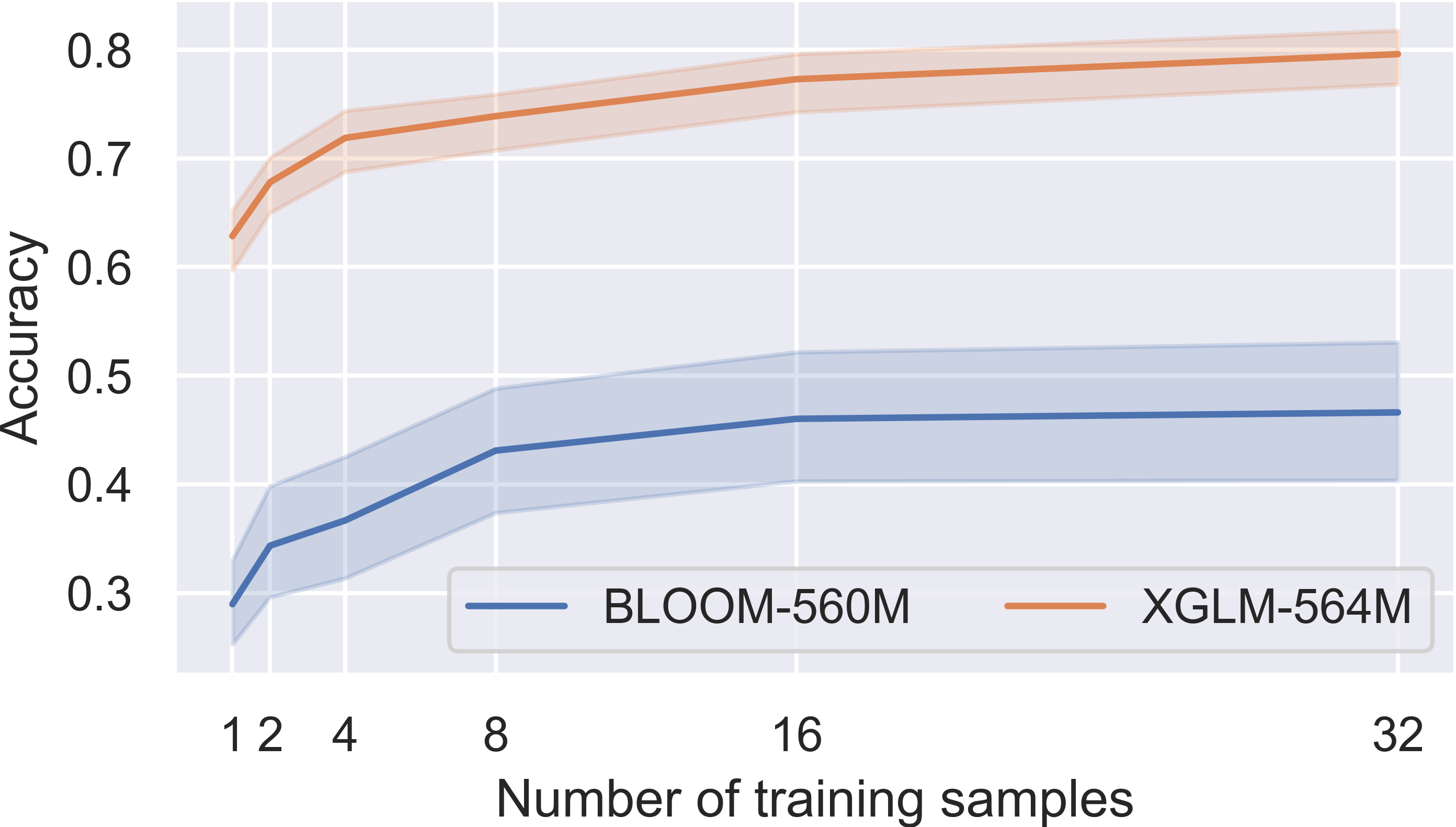}
    \caption{Average cross-lingual transfer performance of SPT with model freezing for different number of training samples per class.}
    \label{fig:n_train}
\end{figure}

Figure \ref{fig:n_train} also shows that, despite the limited number of tunable parameters when freezing all model parameters, additional training samples further boost cross-lingual transfer performance.

\paragraph{Parameter efficiency}
Besides better cross-lingual transfer performance, model freezing during SPT also provides parameter efficiency as fine-tuning is restricted to a number of soft prompt tokens, resulting in only a few thousand parameters in total. This is less than 0.01\% of the parameters fine-tuned in previous studies \citep{zhao_discrete_2021, huang_zero-shot_2022}.

For illustration, the storage requirement for a copy of the XGLM-1.7B model is approximately 3.2 GB, whereas a prompt needs less than 100KB. With respect to training duration, our observations indicate that the time required for training only the soft prompt is less than half compared to when training all model parameters. This benefit becomes even more pronounced when considering the increasing sizes of state-of-the-art models.

\section{Impact of Prompt Length and Reparameterization}

\subsection{Prompt Length} \label{sec:prompt_length}
\begin{figure}[!h]
    \centering
    \includegraphics[width=0.45\textwidth]{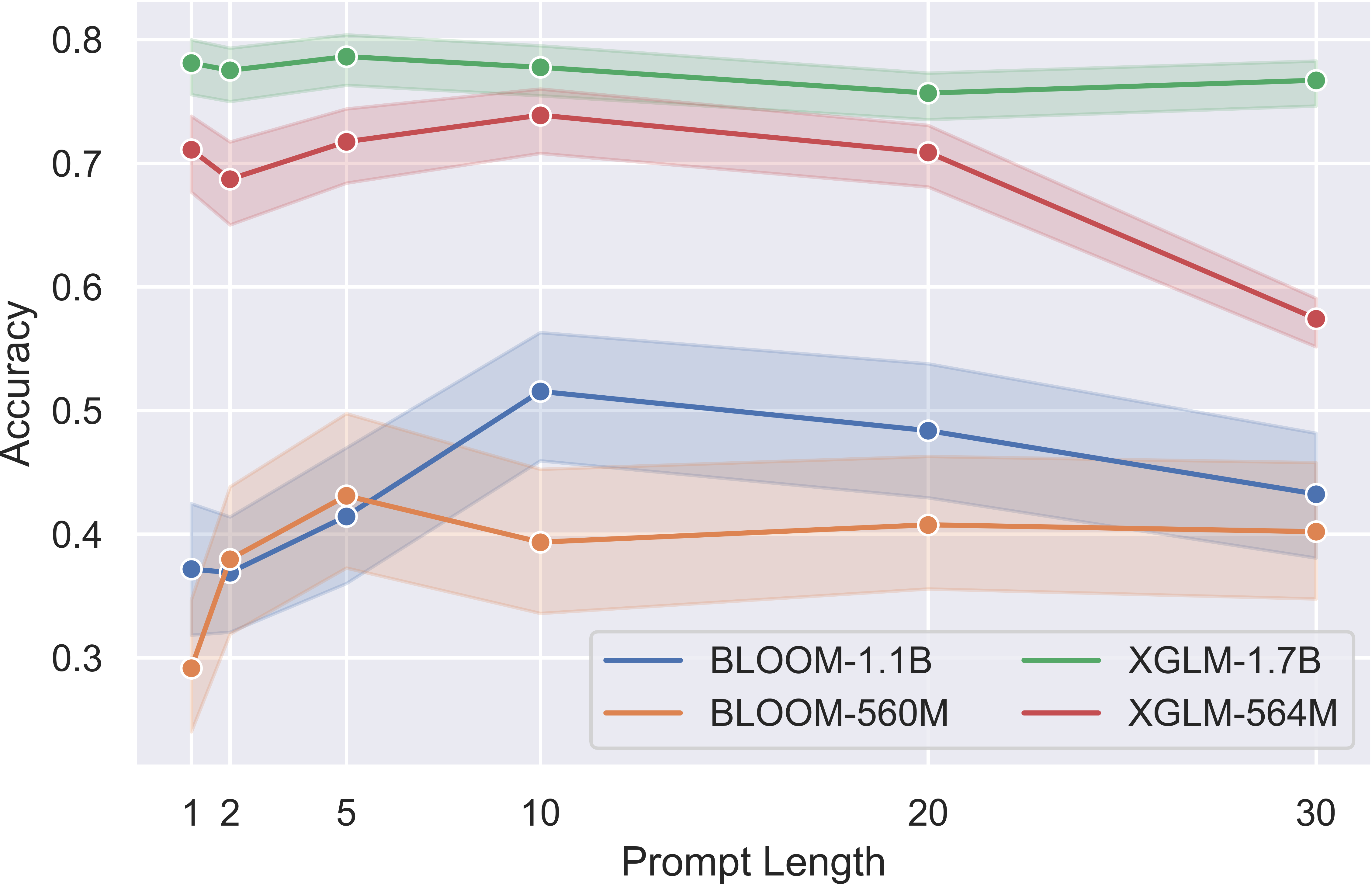}
    \caption{Average cross-lingual transfer performance, measured as accuracy, across different prompt lengths for different models.}
    \label{fig:prompt_length}
\end{figure}

Using the same configuration as described in Section \ref{implementation}, we compare the transfer performance of prompts with different lengths under the 8-shot setting. 
We consider prompt lengths in $\{1,2,5,10,20, 30\}$ and report the results for all 4 models. 
Figure \ref{fig:prompt_length} shows that \textbf{if a soft prompt is too long, cross-lingual transfer performance degrades}.

\subsection{Reparameterization} \label{sec:prompt_reparameterization}
Direct fine-tuning of soft prompt embeddings may lead to unstable training and potentially reduces performance.
To address this issue, previous works have proposed reparameterizing prompt embeddings using different architectures, such as an LSTM \citep{liu_gpt_2021} or MLP \citep{li_prefix-tuning_2021}, which are fine-tuned along with the prompt embeddings. 
\citet{liu_p-tuning_2022} argue that reparameterization can also have negative effects depending on the task or dataset. 

Motivated by this observation, we investigate the effect of reparameterization on cross-lingual transfer performance. 
We adopt the approach proposed by \citet{razdaibiedina_residual_2023}, which uses an MLP with a residual connection and a "bottleneck" layer for reparameterization. We provide further details on this method in Appendix \ref{app:prompt_reparameterization}.

Our analysis reveals that BLOOM is significantly more affected by reparameterization than XGLM (Figure \ref{fig:reparameterization_table} in Appendix \ref{app:prompt_reparameterization}). For both models, the \textbf{impact of reparameterization differs across languages} — being detrimental for some and advantageous for others. Notably, for BLOOM, Atlantic-Congo languages such as Yoruba, Twi, Kinyarwanda, Akan, Fon and Swahili experience the most significant performance decline due to reparameterization, with drops between 24\% to 31\%. Conversely, Indo-Aryan languages like Urdu, Hindi, Bengali, and Nepali, along with Dravidian languages like Malayalam and Tamil see the most significant improvements, with gains of up to 29\%. For XGLM, the outcomes are more balanced. Nonetheless, we observe that the languages that benefit most from reparameterization either use Latin script, such as Haitian, German, and Turkish, or are Dravidian languages such as Telugu and Tamil.

Hence, we recommend that in cross-lingual settings, the decision to use or abstain from reparameterization should not be made uniformly. Instead, it should be tailored based on the specific target languages or language families in consideration.

\section{Discussion}
Previous works on SPT for cross-lingual transfer in few-shot settings suffers from two major drawbacks: 1) fine-tuning all model parameters along with the prompt reduces the computational efficiency of SPT; 2) a bias towards target languages that are linguistically closer to the source language. Our study tackles these issues by showing that by simply keeping model parameters frozen during SPT, we can make progress in addressing both these challenges.

Through our experiments, which covered a wider and more diverse range of languages than prior work on cross-lingual SPT, we observed intriguing effects of non-linguistic variables (such as model freezing, prompt length, and reparameterization) on the transfer performance for individual languages.
Additionally, our results reveal language-specific differences that invite further inquiry into the possibility of tailoring prompts to the target language (e.g., applying prompt reparameterization or not depending on the linguistic distance between the target language family and the source language) rather than using a single prompt for universal transfer across languages.
We believe that our findings will benefit future work on cross-lingual SPT and potentially improve the existing techniques \citep{huang_zero-shot_2022}, becoming more valuable as we adopt larger state-of-the-art models with billion- and trillion-scale parameters \citep{lester_power_2021}.

\section{Conclusion}
The objective of our study was to examine the impact of model freezing on the cross-lingual transfer performance of SPT.
Our results demonstrate that SPT, a method that adjusts less than 0.01\% of parameters compared to full-model fine-tuning, achieves comparable or superior performance for most target languages, particularly for those that are linguistically more distant. Furthermore, we found that shorter prompts enhance SPT's cross-lingual transfer performance, and that some target language families benefit from reparameterization while others are adversely affected by it.
\section*{Limitations}
Our approach enhances transfer performance for several languages, especially those that are linguistically more distant. However, we also notice that it lowers the performance for some languages that are linguistically more similar. This limitation motivates us to pursue future research that aims to achieve balanced performance across languages

Another limitation of our approach is the instability of few-shot fine-tuning, which compromises the robustness of our method's evaluation. To mitigate this issue, we ran all experiments four times with different random seeds and reported the mean and variance of the results. However, we acknowledge that more research is needed to address the challenges of few-shot fine-tuning.
\section*{Ethics Statement}
In this paper, we aim to improve the performance of MLLMs on low-resource languages, which often suffer from a lack of data and attention in NLP research. We believe that this is an important and ethical goal, as it enables NLP advances to benefit a broader range of language communities.

In addition, this paper aims to promote parameter efficiency, which is a crucial factor for reducing the computational and environmental costs of training and fine-tuning state-of-the-art language models. We believe that this aspect will enhance the accessibility and affordability of these models for researchers and practitioners who face computational constraints.

\bibliography{main}

\appendix

\section{Reproducibility} \label{app:reproducibility}
We provide the code used for our experiments here: \href{https://github.com/fredxlpy/cross_lingual_prompt_tuning}{https://github.com/fredxlpy/cross\_lingual\_prompt\_\\tuning}.

\subsection{Dataset}
Our experiments are based on the SIB-200 dataset \citep{adelani_sib-200_2023}. The dataset is based on the FLORES-200 benchmark \citep{nllb_team_no_2022}, and consists of 701 training, 99 validation and 204 test samples in each of the 203 languages. The task is to classify each sample into one of the 7 potential categories: \texttt{science/technology}, \texttt{travel}, \texttt{politics}, \texttt{sports}, \texttt{health}, \texttt{entertainment}, and \texttt{geography}.

\subsection{Models} \label{app:models}
We provide additional information about the models used in our study in Table \ref{tab:models}.

\begin{table}[!h]
    \centering
    \small
    \renewcommand{\arraystretch}{1.5}
    \begin{tabular}
        {|>{\centering\arraybackslash}m{1.2cm}|>{\centering\arraybackslash}m{1cm}|>{\centering\arraybackslash}m{1cm}|>{\centering\arraybackslash}m{1.2cm}|>{\centering\arraybackslash}m{1cm}|}
        \hline
        Model & Layers & Para- meters & Hidden\newline size & Vocab\newline size \\ \hline
        BLOOM-560M & \multirow{6}{*}{24} & 560M & 1.024 & \multirow{3}{*}{250.880} \\ \cline{1-1} \cline{3-4}
        BLOOM-1.1B & & 1.1B & 1.536 & \\ \cline{1-1} \cline{3-5}
        
        XGLM-564M &  & 564M & 1.024 & \multirow{3}{*}{256.008} \\ \cline{1-1} \cline{3-4}
        XGLM-1.7B & & 1.7B & 2.048 & \\ \hline

    \end{tabular}
    \caption{Technical details of the models used in our study.}
    \label{tab:models}
\end{table}

\subsection{Technical Details}

% \usepackage{multirow}
\begin{table}[!h]
\centering
\begin{tabular}{|>{\centering\arraybackslash}m{1.2cm}|>{\centering\arraybackslash}m{1.2cm}|>{\centering\arraybackslash}m{1cm}|>{\centering\arraybackslash}m{1.2cm}|>{\centering\arraybackslash}m{1cm}|}
\hline
 &  & Batch\newline size & Learning\newline rate & Prompt\newline length \\ \hline
\multirow{6}{*}{\parbox{1cm}{\centering w/\\ MF}} & XGLM\newline 564M & \multirow{6}{*}{8} & \multirow{6}{*}{0.1} & \multirow{3}{*}{10} \\ \cline{2-2}
 & XGLM\newline 1.7B &  &  &  \\ \cline{2-2} \cline{5-5} 
 & BLOOM\newline 560M &  &  & 5 \\ \cline{2-2} \cline{5-5} 
 & BLOOM\newline 1.1B &  &  & 10 \\ \hline
\multirow{6}{*}{\parbox{1cm}{\centering w/o\\ MF}} & XGLM\newline 564M & \multirow{6}{*}{8} & \multirow{3}{*}{5e-6} & \multirow{3}{*}{10} \\ \cline{2-2}
 & XGLM\newline 1.7B &  &  &  \\ \cline{2-2} \cline{4-5} 
 & BLOOM\newline 560M &  & \multirow{3}{*}{1e-6} & 5 \\ \cline{2-2} \cline{5-5} 
 & BLOOM\newline 1.1B &  &  & 10 \\ \hline
\end{tabular}
\caption{Hyperparameters used in all of our experiments.} \label{tab:hyperparameters}
\end{table}

We conducted all of our experiments using the \textit{Transformers} library \citep{wolf_transformers_2020}. In a $k$-shot setting, we fine-tune on $k$ samples per class from the English train set and use $\frac{k}{4}$ samples per class for validation. We train all models and prompts for 20 epochs and select the best checkpoint on the development set.
The different hyperparameters used in our experiments are provided in Table \ref{tab:hyperparameters}.

\subsection{Soft Prompt} \label{app:soft_prompt}
We follow the approach of \citet{lester_power_2021} and freeze all model parameters and only fine-tune the soft prompt.

In order to map the tokens predicted by the model to the respective class, we define a verbalizer $F:T \rightarrow C$, where $T=\{t_1,\ldots,t_K\}$ is a subset of the model's vocabulary $V$ and $C=\{1,\ldots,K\}$ are the respective classes.

We append a prompt $p=\{p_1,\ldots,p_m\}$ to an input sequence $x=\{x_1,\ldots,x_n\}$ and pass $\{x_1,\ldots,x_n,p_1,\ldots,p_m\}$ through the autoregressive language model which outputs the logits for the next token in the input sequence $\{l_1,\ldots,l_{|V|}\}$.

The predicted token is then $F\left(argmax_{i \in T} l_i\right)$

\subsection{Computing Resources}
We conduct all our experiments on 4 A100 40GB GPUs, using 4 different random seeds, in parallel. All experiments could be run in a few hours.

\section{Language Distance Metrics} \label{app:lang_dist}
We consider six types of lang2vec\footnote{\url{https://github.com/antonisa/lang2vec}} \citep{littell_uriel_2017} distances:

\begin{itemize}
\item \textbf{Syntactic Distance} (SYN) captures the similarity of syntactic structures across languages. It is computed as the cosine distance between syntax feature vectors, which are derived from the World Atlas of Language Structures\footnote{\url{https://wals.info}} (WALS) \citep{dryer_wals_2013}, Syntactic Structures of World Languages\footnote{\url{http://sswl.railsplayground.net/}} (SSWL) \citep{collins_syntactic_2011} and Ethnologue\footnote{\url{https://www.ethnologue.com/}} \citep{lewis_ethnologue_2015}.
\item \textbf{Geographic Distance} (GEO) reflects the spatial proximity of languages. It is calculated as the shortest distance between two languages on the surface of the earth’s sphere (i.e., orthodromic distance).
\item \textbf{Inventory Distance} (INV) measures the difference in sound inventories across languages. It is computed as the cosine distance between inventory feature vectors, which are obtained from the PHOIBLE\footnote{\url{https://phoible.org/}} database \citep{moran_phoible_2019}.
\item \textbf{Genetic Distance} (GEN) indicates the historical relatedness of languages. It is based on the Glottolog\footnote{\url{https://glottolog.org}} \citep{hammarstrom_glottolog_2015} tree of language families and is obtained by computing the distance between two languages in the tree. \item \textbf{Phonological Distance} (PHON) captures the similarity of sound patterns across languages. It is computed as the cosine distance between phonological feature vectors, which are sourced from WALS and Ethnologue.
\item \textbf{Featural Distance} (FEA) is the cosine distance between feature vectors from a combination of the 5 above-listed linguistic features.
\end{itemize}

The values for each distance type range from 0 to 1, where 0 indicates the minimum distance and 1 indicates the maximum distance.

\section{Prompt Reparameterization} \label{app:prompt_reparameterization}
\begin{figure}
    \centering
    \includegraphics[width=0.45\textwidth]{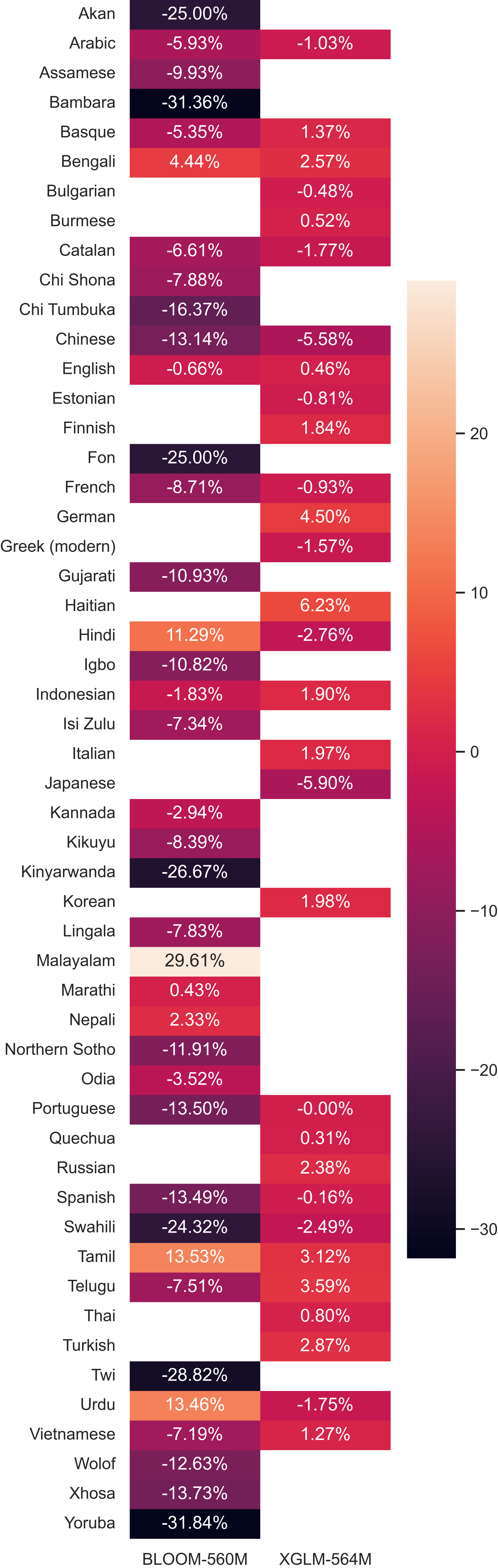}
    \caption{Impact of reparameterization (expressed in \%) on the cross-lingual transfer performance of BLOOM-560M and XGLM-564M for different target languages.} 
    \label{fig:reparameterization_table}
\end{figure}

We follow the residual reparameterization method of \citet{razdaibiedina_residual_2023} to examine the impact of soft prompt reparameterization. This method employs a multi-layer perceptron (MLP) architecture for the reparameterization network, which consists of a \textit{down-projection} layer and an \textit{up-projection} layer with parameter $W_{down} \in \mathbb{R}^{d \times m}$ and $W_{up} \in \mathbb{R}^{m \times d}$ respectively, where $d$ denotes the model embedding size and $m$ denotes the hidden representation dimension between both layers (\textit{bottleneck size}). A ReLU layer is applied to the hidden representation, and a normalization layer is applied to the output of the \textit{up-projection} layer before summing it with the initial input embedding via a residual connection.
We fine-tune the soft prompt and its reparameterization network with a bottleneck size of 500 for BLOOM-560M and 200 for XGLM-564M and report the impact of reparameterization across all target languages in Figure \ref{fig:reparameterization_table}. Except for the reparameterization, we adopt the same implementation settings as described in Section \ref{sec:experimental_setup}.

\section{Full Results}\label{app:results}
The full results discussed in Section \ref{sec:results} are provided in Table \ref{table:results}.

\begin{table*}[]
\small
\centering
\setlength{\tabcolsep}{8pt} % Default value: 6pt
\renewcommand{\arraystretch}{1.2} % Default value: 1
\begin{tabular}{c|cc|cc|cc|cc}
\textbf{} & \multicolumn{2}{c|}{\textbf{BLOOM-560M}} & \multicolumn{2}{c|}{\textbf{BLOOM-1.1B}} & \multicolumn{2}{c|}{\textbf{XGLM-564M}} & \multicolumn{2}{c|}{\textbf{XGLM-1.7B}} \\
\textbf{Language} & \textbf{w/o MF} & \textbf{w/ MF} & \textbf{w/o MF} & \textbf{w/ MF} & \textbf{w/o MF} & \textbf{w/ MF} & \textbf{w/o MF} & \multicolumn{1}{c|}{\textbf{w/ MF}} \\ \hline
Akan & $\text{22,18}_\text{9,67}$ & \textbf{$\text{34,80}_\text{6,33}$} & $\text{19,36}_\text{6,52}$ & \textbf{$\text{35,05}_\text{0,85}$} & - & - & - & - \\
Arabic & $\text{55,51}_\text{3,56}$ & \textbf{$\text{70,22}_\text{1,62}$} & $\text{42,03}_\text{11,9}$ & \textbf{$\text{63,48}_\text{3,50}$} & $\text{57,60}_\text{3,34}$ & \textbf{$\text{71,69}_\text{1,89}$} & $\text{74,75}_\text{1,67}$ & \textbf{$\text{78,68}_\text{5,49}$} \\
Assamese & $\text{27,45}_\text{7,99}$ & \textbf{$\text{37,01}_\text{4,85}$} & $\text{29,41}_\text{9,66}$ & \textbf{$\text{53,06}_\text{4,92}$} & - & - & - & - \\
Bambara & $\text{16,67}_\text{3,63}$ & \textbf{$\text{26,96}_\text{8,74}$} & $\text{17,03}_\text{5,94}$ & \textbf{$\text{29,17}_\text{3,68}$} & - & - & - & - \\
Basque & $\text{43,50}_\text{12,5}$ & \textbf{$\text{61,89}_\text{1,90}$} & $\text{38,73}_\text{7,69}$ & \textbf{$\text{63,97}_\text{13,0}$} & $\text{67,40}_\text{1,30}$ & \textbf{$\text{71,32}_\text{2,70}$} & $\text{71,08}_\text{3,07}$ & \textbf{$\text{72,43}_\text{6,64}$} \\
Bengali & $\text{56,62}_\text{4,48}$ & \textbf{$\text{60,78}_\text{2,41}$} & $\text{46,69}_\text{12,2}$ & \textbf{$\text{71,81}_\text{2,90}$} & $\text{68,14}_\text{3,51}$ & \textbf{$\text{71,45}_\text{4,24}$} & $\text{71,57}_\text{3,05}$ & \textbf{$\text{76,23}_\text{5,22}$} \\
Bulgarian & - & - & - & - & $\text{72,79}_\text{5,13}$ & \textbf{$\text{77,33}_\text{2,45}$} & $\text{78,92}_\text{2,40}$ & \textbf{$\text{81,37}_\text{4,33}$} \\
Burmese & - & - & - & - & $\text{63,60}_\text{6,06}$ & \textbf{$\text{71,20}_\text{3,38}$} & $\text{72,67}_\text{3,03}$ & \textbf{$\text{73,41}_\text{7,79}$} \\
Catalan & $\text{63,48}_\text{13,1}$ & \textbf{$\text{72,30}_\text{2,28}$} & $\text{48,77}_\text{7,93}$ & \textbf{$\text{73,77}_\text{4,03}$} & $\text{68,50}_\text{7,52}$ & \textbf{$\text{76,35}_\text{3,03}$} & $\text{77,33}_\text{4,01}$ & \textbf{$\text{79,04}_\text{4,13}$} \\
Chi Shona & $\text{19,98}_\text{4,79}$ & \textbf{$\text{24,88}_\text{2,67}$} & $\text{17,89}_\text{5,69}$ & \textbf{$\text{31,00}_\text{3,95}$} & - & - & - & - \\
Chi Tumbuka & $\text{20,34}_\text{4,54}$ & \textbf{$\text{27,70}_\text{2,55}$} & $\text{18,14}_\text{4,95}$ & \textbf{$\text{33,70}_\text{4,62}$} & - & - & - & - \\
Chinese & $\text{60,54}_\text{11,1}$ & \textbf{$\text{73,65}_\text{6,47}$} & $\text{47,30}_\text{13,9}$ & \textbf{$\text{72,43}_\text{3,36}$} & $\text{59,93}_\text{8,33}$ & \textbf{$\text{79,04}_\text{1,85}$} & $\text{77,94}_\text{5,08}$ & \textbf{$\text{81,74}_\text{4,28}$} \\
English & \textbf{$\text{75,00}_\text{5,87}$} & $\text{74,63}_\text{2,09}$ & $\text{69,36}_\text{2,67}$ & \textbf{$\text{75,12}_\text{2,90}$} & $\text{78,68}_\text{1,67}$ & \textbf{$\text{79,90}_\text{2,62}$} & $\text{80,88}_\text{2,94}$ & \textbf{$\text{82,84}_\text{5,41}$} \\
Estonian & - & - & - & - & $\text{72,30}_\text{3,24}$ & \textbf{$\text{75,86}_\text{3,13}$} & $\text{76,35}_\text{1,76}$ & \textbf{$\text{81,13}_\text{5,78}$} \\
Finnish & - & - & - & - & $\text{76,72}_\text{1,81}$ & \textbf{$\text{79,90}_\text{1,44}$} & $\text{79,78}_\text{1,76}$ & \textbf{$\text{82,35}_\text{5,92}$} \\
Fon & $\text{19,36}_\text{10,0}$ & \textbf{$\text{25,49}_\text{7,88}$} & $\text{13,97}_\text{3,98}$ & \textbf{$\text{26,84}_\text{5,51}$} & - & - & - & - \\
French & $\text{69,61}_\text{6,52}$ & \textbf{$\text{73,16}_\text{1,89}$} & $\text{57,23}_\text{6,29}$ & \textbf{$\text{72,92}_\text{5,51}$} & $\text{71,94}_\text{4,26}$ & \textbf{$\text{79,29}_\text{2,98}$} & $\text{79,04}_\text{5,48}$ & \textbf{$\text{79,90}_\text{2,80}$} \\
German & - & - & - & - & $\text{71,57}_\text{7,19}$ & \textbf{$\text{76,23}_\text{4,67}$} & $\text{81,62}_\text{5,04}$ & $\text{81,62}_\text{5,79}$ \\
Greek (modern) & - & - & - & - & $\text{73,90}_\text{3,47}$ & \textbf{$\text{78,19}_\text{2,93}$} & $\text{80,27}_\text{2,70}$ & \textbf{$\text{82,97}_\text{5,11}$} \\
Gujarati & \textbf{$\text{41,79}_\text{7,92}$} & $\text{37,01}_\text{9,35}$ & $\text{27,08}_\text{7,85}$ & \textbf{$\text{54,29}_\text{10,3}$} & - & - & - & - \\
Haitian & - & - & - & - & $\text{65,44}_\text{1,30}$ & \textbf{$\text{68,87}_\text{2,55}$} & $\text{74,39}_\text{1,72}$ & \textbf{$\text{74,75}_\text{6,70}$} \\
Hindi & $\text{42,52}_\text{4,28}$ & \textbf{$\text{45,59}_\text{4,47}$} & $\text{50,12}_\text{10,0}$ & \textbf{$\text{64,95}_\text{2,85}$} & $\text{74,14}_\text{3,28}$ & \textbf{$\text{75,37}_\text{2,41}$} & $\text{75,74}_\text{2,95}$ & \textbf{$\text{78,19}_\text{4,88}$} \\
Igbo & $\text{18,50}_\text{1,57}$ & \textbf{$\text{23,77}_\text{6,42}$} & $\text{15,20}_\text{4,85}$ & \textbf{$\text{27,70}_\text{4,57}$} & - & - & - & - \\
Indonesian & $\text{49,26}_\text{2,55}$ & \textbf{$\text{66,91}_\text{1,86}$} & $\text{49,14}_\text{11,9}$ & \textbf{$\text{68,75}_\text{3,38}$} & $\text{73,90}_\text{1,29}$ & \textbf{$\text{77,57}_\text{2,45}$} & $\text{77,21}_\text{2,48}$ & \textbf{$\text{79,90}_\text{5,34}$} \\
Isi Zulu & $\text{19,24}_\text{6,01}$ & \textbf{$\text{21,69}_\text{2,72}$} & $\text{15,69}_\text{5,98}$ & \textbf{$\text{29,66}_\text{2,48}$} & - & - & - & - \\
Italian & - & - & - & - & $\text{73,41}_\text{4,82}$ & \textbf{$\text{74,75}_\text{1,52}$} & $\text{78,43}_\text{4,95}$ & \textbf{$\text{80,02}_\text{5,52}$} \\
Japanese & - & - & - & - & $\text{54,29}_\text{5,98}$ & \textbf{$\text{76,84}_\text{3,89}$} & \textbf{$\text{80,64}_\text{1,47}$} & $\text{77,94}_\text{4,65}$ \\
Kannada & $\text{22,30}_\text{8,24}$ & \textbf{$\text{25,00}_\text{8,46}$} & $\text{22,92}_\text{3,85}$ & \textbf{$\text{55,76}_\text{7,93}$} & - & - & - & - \\
Kikuyu & $\text{28,19}_\text{8,36}$ & \textbf{$\text{35,05}_\text{2,42}$} & $\text{19,49}_\text{4,44}$ & \textbf{$\text{33,70}_\text{3,81}$} & - & - & - & - \\
Kinyarwanda & $\text{19,00}_\text{3,21}$ & \textbf{$\text{25,74}_\text{6,26}$} & $\text{15,69}_\text{3,80}$ & \textbf{$\text{30,39}_\text{4,33}$} & - & - & - & - \\
Korean & - & - & - & - & $\text{73,77}_\text{1,67}$ & \textbf{$\text{74,26}_\text{2,28}$} & $\text{74,75}_\text{4,46}$ & \textbf{$\text{77,45}_\text{5,41}$} \\
Lingala & $\text{23,90}_\text{3,85}$ & \textbf{$\text{28,19}_\text{4,74}$} & $\text{21,69}_\text{8,43}$ & \textbf{$\text{36,15}_\text{3,29}$} & - & - & - & - \\
Malayalam & \textbf{$\text{23,53}_\text{11,1}$} & $\text{21,94}_\text{7,56}$ & $\text{30,39}_\text{9,95}$ & \textbf{$\text{59,93}_\text{4,17}$} & - & - & - & - \\
Marathi & \textbf{$\text{34,68}_\text{11,1}$} & $\text{28,31}_\text{5,83}$ & $\text{29,78}_\text{6,21}$ & \textbf{$\text{60,05}_\text{4,41}$} & - & - & - & - \\
Nepali & $\text{30,15}_\text{6,99}$ & \textbf{$\text{42,03}_\text{6,95}$} & $\text{36,76}_\text{13,3}$ & \textbf{$\text{67,03}_\text{6,25}$} & - & - & - & - \\
Northern Sotho & $\text{20,59}_\text{6,62}$ & \textbf{$\text{28,80}_\text{0,47}$} & $\text{18,38}_\text{4,09}$ & \textbf{$\text{33,82}_\text{2,40}$} & - & - & - & - \\
Odia & \textbf{$\text{34,80}_\text{7,64}$} & $\text{31,37}_\text{6,62}$ & $\text{25,00}_\text{5,25}$ & \textbf{$\text{47,06}_\text{9,22}$} & - & - & - & - \\
Portuguese & $\text{66,67}_\text{5,02}$ & \textbf{$\text{75,37}_\text{3,19}$} & $\text{53,19}_\text{5,69}$ & \textbf{$\text{73,77}_\text{2,17}$} & $\text{74,26}_\text{1,90}$ & \textbf{$\text{79,53}_\text{1,09}$} & $\text{80,15}_\text{1,98}$ & \textbf{$\text{82,48}_\text{3,95}$} \\
Quechua & - & - & - & - & $\text{35,66}_\text{8,69}$ & \textbf{$\text{39,71}_\text{2,23}$} & $\text{49,88}_\text{4,84}$ & \textbf{$\text{51,59}_\text{6,37}$} \\
Russian & - & - & - & - & $\text{76,96}_\text{3,23}$ & \textbf{$\text{77,21}_\text{1,98}$} & $\text{78,19}_\text{3,43}$ & \textbf{$\text{80,27}_\text{4,30}$} \\
Spanish & $\text{63,36}_\text{8,94}$ & \textbf{$\text{72,67}_\text{0,47}$} & $\text{46,69}_\text{9,79}$ & \textbf{$\text{73,65}_\text{5,11}$} & $\text{71,45}_\text{0,74}$ & \textbf{$\text{76,47}_\text{2,30}$} & $\text{77,33}_\text{3,63}$ & \textbf{$\text{79,78}_\text{4,84}$} \\
Swahili & $\text{35,05}_\text{7,95}$ & \textbf{$\text{49,88}_\text{6,02}$} & $\text{25,12}_\text{6,22}$ & \textbf{$\text{49,75}_\text{7,40}$} & $\text{61,40}_\text{8,61}$ & \textbf{$\text{69,00}_\text{2,84}$} & \textbf{$\text{73,77}_\text{2,45}$} & $\text{72,79}_\text{7,91}$ \\
Tamil & $\text{44,85}_\text{9,58}$ & \textbf{$\text{50,74}_\text{4,09}$} & $\text{34,44}_\text{13,1}$ & \textbf{$\text{67,40}_\text{4,71}$} & $\text{68,75}_\text{5,68}$ & \textbf{$\text{70,59}_\text{2,12}$} & $\text{73,90}_\text{1,01}$ & \textbf{$\text{75,86}_\text{7,91}$} \\
Telugu & $\text{24,51}_\text{3,94}$ & \textbf{$\text{31,00}_\text{6,71}$} & $\text{26,96}_\text{1,20}$ & \textbf{$\text{66,05}_\text{7,13}$} & $\text{62,75}_\text{3,33}$ & \textbf{$\text{68,26}_\text{5,15}$} & $\text{74,14}_\text{3,76}$ & \textbf{$\text{76,23}_\text{6,46}$} \\
Thai & - & - & - & - & $\text{67,77}_\text{6,42}$ & \textbf{$\text{76,35}_\text{1,16}$} & \textbf{$\text{79,53}_\text{1,72}$} & $\text{77,33}_\text{5,02}$ \\
Turkish & - & - & - & - & $\text{73,16}_\text{2,84}$ & \textbf{$\text{76,96}_\text{3,18}$} & $\text{74,63}_\text{4,30}$ & \textbf{$\text{79,17}_\text{5,89}$} \\
Twi & $\text{23,41}_\text{9,5}$ & \textbf{$\text{35,29}_\text{6,64}$} & $\text{18,75}_\text{6,83}$ & \textbf{$\text{36,52}_\text{3,32}$} & - & - & - & - \\
Urdu & $\text{42,28}_\text{6,67}$ & \textbf{$\text{44,61}_\text{8,95}$} & $\text{31,74}_\text{8,12}$ & \textbf{$\text{48,41}_\text{9,35}$} & $\text{54,90}_\text{8,37}$ & \textbf{$\text{70,10}_\text{2,86}$} & $\text{70,10}_\text{3,12}$ & \textbf{$\text{75,25}_\text{5,69}$} \\
Vietnamese & $\text{46,08}_\text{19,4}$ & \textbf{$\text{68,14}_\text{7,21}$} & $\text{43,87}_\text{7,49}$ & \textbf{$\text{64,58}_\text{3,76}$} & $\text{70,71}_\text{3,06}$ & \textbf{$\text{76,96}_\text{3,18}$} & $\text{78,31}_\text{3,63}$ & \textbf{$\text{79,90}_\text{7,30}$} \\
Wolof & $\text{25,49}_\text{7,88}$ & \textbf{$\text{34,93}_\text{4,17}$} & $\text{21,81}_\text{9,77}$ & \textbf{$\text{41,42}_\text{4,64}$} & - & - & - & - \\
Xhosa & $\text{21,94}_\text{7,35}$ & \textbf{$\text{28,55}_\text{1,23}$} & $\text{15,32}_\text{5,51}$ & \textbf{$\text{32,23}_\text{6,14}$} & - & - & - & - \\
Yoruba & $\text{13,36}_\text{1,62}$ & \textbf{$\text{21,94}_\text{9,49}$} & $\text{16,30}_\text{2,28}$ & \textbf{$\text{33,21}_\text{4,76}$} & - & - & - & - \\ \hline
\end{tabular}
\caption{Cross-lingual transfer results, reported as accuracy, along with standard deviation across 4 runs, after 8-shot soft prompt tuning (SPT) in English, with and without model freezing (MF).} \label{table:results}
\end{table*}

\end{document}